\documentclass[twoside]{article}

\usepackage[T1]{fontenc}

\usepackage[utf8]{inputenc} 
\usepackage{tgtermes}
\usepackage{amssymb}
\usepackage{amsfonts}

\usepackage{tgtermes}
\usepackage{amsmath}

\usepackage{scalefnt,letltxmacro}
\LetLtxMacro{\oldtextsc}{\textsc}
\renewcommand{\textsc}[1]{\oldtextsc{\scalefont{1.10}#1}}
\usepackage[scaled=0.92]{PTSans}

\usepackage[usenames,dvipsnames]{xcolor}
\definecolor{shadecolor}{gray}{0.9}

\usepackage[final,expansion=alltext]{microtype}
\usepackage[english]{babel}
\usepackage[parfill]{parskip}
\usepackage{afterpage}
\usepackage{framed}
\usepackage{nicefrac}
\usepackage{bm}
\usepackage{fixmath}

\usepackage[colorinlistoftodos,
           prependcaption,
           textsize=small,
           backgroundcolor=yellow,
           linecolor=lightgray,
           bordercolor=lightgray]{todonotes}

\usepackage{lineno}

\usepackage{ragged2e}

\DeclareRobustCommand{\parhead}[1]{\textbf{#1}~}

\usepackage{graphicx}
\usepackage[labelfont=bf]{caption}
\usepackage[format=hang]{subcaption}

\usepackage{booktabs}
\usepackage{arydshln} 
\usepackage{natbib}

\usepackage{listings}
\usepackage{fancyvrb}
\fvset{fontsize=\normalsize}

\usepackage[colorlinks,linktoc=all]{hyperref}
\usepackage[all]{hypcap}
\hypersetup{citecolor=Violet}
\hypersetup{linkcolor=black}
\hypersetup{urlcolor=MidnightBlue}
\usepackage{url}

\usepackage[acronym,smallcaps,nowarn]{glossaries}

\usepackage{listings}
\lstdefinestyle{alp_style}{
    commentstyle=\color{OliveGreen},
    numberstyle=\tiny\color{black!60},
    stringstyle=\color{BrickRed},
    basicstyle=\ttfamily\scriptsize,
    breakatwhitespace=false,
    breaklines=true,
    captionpos=b,
    keepspaces=true,
    numbers=none,
    numbersep=5pt,
    showspaces=false,
    showstringspaces=false,
    showtabs=false,
    tabsize=2
}
\newcommand{\g}{\,|\,}
\renewcommand{\gg}{\,\|\,}

\usepackage{bbold}

\newtheorem{thm}{Theorem}

\usepackage{mathtools}

\newcommand{\mbh}{\mathbold{h}}

\newcommand{\mbx}{\mathbold{x}}

\newcommand{\mbz}{\mathbold{z}}

\newcommand{\mbI}{\mathbold{I}}

\newcommand{\mbzero}{\boldsymbol{0}}

 \newacronym{BBVI}{bbvi}{black box variational inference}
\newacronym{ELBO}{elbo}{evidence lower bound}
\newacronym{VI}{vi}{variational inference}
\newacronym{KL}{kl}{Kullback-Leibler}
\newacronym{MCMC}{mcmc}{Markov chain Monte Carlo}
\newacronym{klVI}{klvi}{$\operatorname{KL}(q || p)$ variational inference}
\newacronym{VAE}{vae}{variational autoencoder}
\newacronym{SAVAE}{sa-vae}{semi-amortized variational autoencoder}
\newacronym{skipVAE}{skip-vae}{Skip Variational Autoencoder}
\newacronym{SkipSAVAE}{skip-sa-vae}{Skip Variational Autoencoder}
\newacronym{LM}{language model}{Language Model}
\newacronym{GSM}{gsm}{generative skip model}
\newacronym{MLP}{mlp}{multilayer perceptron} 
\usepackage{bbm}
\DeclareMathOperator*{\kl}{KL}

\newcommand{\xvec}{\mathbf{x}}
\newcommand{\zvec}{\mathbf{z}}

\newtheorem{definition}{Definition}

\usepackage{multirow} 
 \usepackage[accepted]{aistats2019}

\hypersetup{citecolor=Violet}
\hypersetup{linkcolor=black}
\hypersetup{urlcolor=MidnightBlue}

\begin{document}

\twocolumn[

\aistatstitle{Avoiding Latent Variable Collapse with Generative Skip Models}

\aistatsauthor{ Adji B. Dieng \And Yoon Kim \And Alexander M. Rush \And David M. Blei }
\aistatsaddress{ Columbia University \And  Harvard University \And Harvard University \And Columbia University } ]
\begin{abstract}
\vskip 0.1in
\Glspl{VAE} learn distributions of high-dimensional data.~They model data with a deep latent-variable model and then fit the model by maximizing a lower bound of the log marginal likelihood.~\glspl{VAE} can capture complex distributions, but they can also suffer from an issue known as "latent variable collapse," especially if the likelihood model is powerful. Specifically, the lower bound involves an approximate posterior of the latent variables; this posterior "collapses" when it is set equal to the prior, i.e., when the approximate posterior is independent of the data.~While \glspl{VAE} learn good generative models, latent variable collapse prevents them from learning useful representations.~In this paper, we propose a simple new way to avoid latent variable collapse by including skip connections in our generative model; these connections enforce strong links between the latent variables and the likelihood function.~We study generative skip models both theoretically and empirically.~Theoretically, we prove that skip models increase the mutual information between the observations and the inferred latent variables.~Empirically, we study images (MNIST and Omniglot) and text (Yahoo). Compared to existing VAE architectures, we show that generative skip models maintain similar predictive performance but lead to less collapse and provide more meaningful representations of the data.
\end{abstract}

\glsresetall
\section{Introduction}\label{sec:introduction}

Unsupervised representation learning aims to find good low-dimensional representations of high-dimensional data.  One powerful method for representation learning is the \gls{VAE}~\citep{kingma2013auto,rezende2014stochastic}. \gls{VAE}s have been studied for text analysis~\citep{bowman2015generating, miao2016neural, dieng2016topicrnn, guu2017generating, xu2018spherical},
collaborative filtering~\citep{liang2018variational}, dialog modeling~\citep{zhao2018unsupervised}, image
analysis~\citep{chen2016variational, van2017neural}, and many other applications.

A \gls{VAE} binds together modeling and inference. The model is a deep generative model, which defines a joint distribution of latent variables $\mbz$ and observations $\mbx$,
\begin{align*}
  p_{\theta}(\mbx, \mbz)
  &= p_{\theta}(\mbx \g \mbz) p(\mbz).
\end{align*}
A typical \gls{VAE} uses a spherical Gaussian prior
$p(\mbz) = \mathcal{N}(\mbzero, \mbI)$ and a likelihood parameterized by a deep neural network. Specifically, the likelihood of observation $\mbx_i$ is an exponential family whose natural parameter $\eta(\mbz_i; \theta)$ is a deep network with the latent representation $\mbz_i$ as input. Inference in VAEs is performed with variational methods.
\gls{VAE}s are powerful, but they can suffer from a phenomenon known as \textit{latent variable collapse}~\citep{bowman2015generating, hoffman2016elbo, sonderby2016train, kingma2016improved, chen2016variational, zhao2017towards,yeung2017tackling, alemi2018fixing}, in which the variational posterior collapses to the prior. When this phenomenon occurs, the \gls{VAE} can learn a good generative model of the data but still fail to learn good representations of the individual data points. We propose a new way to avoid this issue.

Ideally the parameters of the deep generative model
should be fit by maximizing the marginal likelihood of the observations,
\begin{align}
  \label{eq:marginal-likelihood}
\theta^*
  &= \arg \max_{\theta}
    \sum_{i=1}^{N} \log \int p_{\theta}(\mbx_i, \mbz_i) d\mbz.
\end{align}
However each term of this objective contains an intractable integral. To this end, \glspl{VAE} rely on amortized variational inference to approximate the posterior distribution. First posit a variational approximation $q_{\phi}(\mbz \g \mbx_i)$. This is an \textit{amortized} family, a distribution over latent variables $\mbz_i$ that takes the observation $\mbx_i$ as input and uses a deep neural network to produce variational parameters.  Using this family, the \gls{VAE} optimizes the \gls{ELBO},
\begin{align}
	\gls{ELBO} &= 
	\sum_{i=1}^{N} E_{q_{\phi}(\mbz_i \g \mbx_i)} \left[
    \log p_{\theta}(\mbx_i \g \mbz_i)\right] \nonumber \\
    &- \text{KL}(q_{\phi}(\mbz_i \g \mbx_i) \gg p(\mbz_i))\label{eq:elbo1}
    .
\end{align}
The \gls{ELBO} is a lower bound on the log marginal likelihood of Eq.\nobreakspace \ref {eq:marginal-likelihood}. Thus the \gls{VAE} optimizes Eq.\nobreakspace \ref {eq:elbo1} with respect to both the generative model parameters $\theta$ and the variational neural network parameters $\phi$.  Fitting $\theta$ finds a good model; fitting $\phi$ finds a neural network that produces good approximate posteriors.

This method is theoretically sound. Empirically, however, fitting Eq.\nobreakspace \ref {eq:elbo1} often leads to a degenerate solution where
\begin{align*}
  q_{\phi}(\mbz_i \g \mbx_i) &\approx p(\mbz_i),
\end{align*}
i.e. the variational ``posterior'' does not depend on the data; this is known as latent variable collapse. When the posterior collapses, $\mbz$ and $\mbx$ are essentially independent and consequently posterior estimates of the latent variable $\mbz$ do not represent faithful summaries of their data $\mbx$---the \gls{VAE} has not learned good representations. This issue is especially a problem when the likelihood $p_{\theta}(\mbx_i \g \mbz_i)$ has high capacity \citep{bowman2015generating, sonderby2016train, kingma2016improved, chen2016variational, zhao2017towards,yeung2017tackling}.  

We propose a new method to alleviate latent variable collapse. The idea is to add skip connections in the deep generative model that parameterizes the likelihood function. Skip connections attach the latent input $\mbz_i$ to multiple layers in the model's neural network. The resulting \emph{generative skip model} is at least as expressive as the original deep generative model, but it forces the likelihood to maintain a strong connection between the latent variables $\mbz_i$ and the observations $\mbx_i$. Consequently, as we show, posterior estimates of $\mbz_i$ provide good representations of the data.

\glspl{VAE} with generative skip models---which we call \glspl{skipVAE}---produce both good generative models and good representations. Section\nobreakspace \ref {sec:empirical} studies the traditional \gls{VAE}~\citep{kingma2013auto,rezende2014stochastic} with PixelCNN/LSTM generative models analyzing both text and image datasets. For similar levels of model performance, as measured by the approximate likelihood, \glspl{skipVAE} promote more dependence between $\mbx$ and $\mbz$ as measured by mutual information and other metrics in Section\nobreakspace \ref {sec:empirical}. Moreover, the advantages of \glspl{skipVAE} increase as the generative model gets deeper. 

Generative skip models can be used in concert with other techniques. For example~Section\nobreakspace \ref {sec:empirical} also studies generative skip models with the \gls{SAVAE}~\citep{kim2018semi}\footnote{resulting in the \acrshort{SkipSAVAE}}, which have also been shown to mitigate posterior collapse. When used with the \gls{SAVAE}, generative skip models further improve the learned representations.

\parhead{Related Work.}
Skip connections are widely used in deep learning, for example, in designing residual, highway, and attention networks~\citep{fukushima1988neocognitron,he2016identity,srivastava2015highway,bahdanau2014neural}. They have not been studied for alleviating latent variable collapse in \glspl{VAE}. 

Many papers discuss latent variable collapse \citep{bowman2015generating, hoffman2016elbo, sonderby2016train, kingma2016improved, chen2016variational, zhao2017towards,yeung2017tackling, alemi2018fixing}. To address it, the most common heuristic is to anneal the $\kl$ term in the \gls{VAE} objective~\citep{bowman2015generating,sonderby2016train}.

Several other solutions have also been proposed. One approach is to handicap the training of the generative model \citep{bowman2015generating} or weaken its capacity \citep{gulrajani2016pixelvae,Yang2017}, effectively encouraging better representations by limiting the generative model. Another approach 
replaces the simple spherical Gaussian prior with more sophisticated priors. For example \cite{van2017neural} and \cite{ tomczak2017vae} propose parametric priors, which are learned along with the generative model. Still another approach uses richer variational distributions \citep{rezende2015variational}. In another thread of research, \cite{makhzani2015adversarial} and \cite{mescheder2017adversarial} replace the $\kl$ regularization term in the \gls{VAE} objective with adversarial regularizers. \cite{higgins2016beta} dampen the effect of the $\kl$ regularization term with Lagrange multipliers. Finally, one can appeal to new inference algorithms. For example \cite{hoffman2017learning} uses \gls{MCMC} instead of variational inference and \cite{kim2018semi} uses stochastic variational inference, initialized with the variational neural network parameters, to iteratively refine the variational distribution. 

A very recent approach to address posterior collapse relies on ideas from directional statistics. \cite{guu2017generating} and \cite{xu2018spherical} use the Von Mises-Fisher distribution for both the prior and the variational posterior and fixing the dispersion parameter of the Von Mises-Fisher distribution to make the $\kl$ term in the \gls{ELBO} constant~. We note this practice might result in less expressive approximate posteriors. 

The generative skip models we propose differ from all of these strategies and can potentially complement them. They modify the generative model of the \gls{VAE} using skip connections to enforce dependence between the observations and their latent variables. They can be used with any prior and variational distributions. 
\begin{figure*}[htp]
  \centering
  {\includegraphics[scale=0.25]{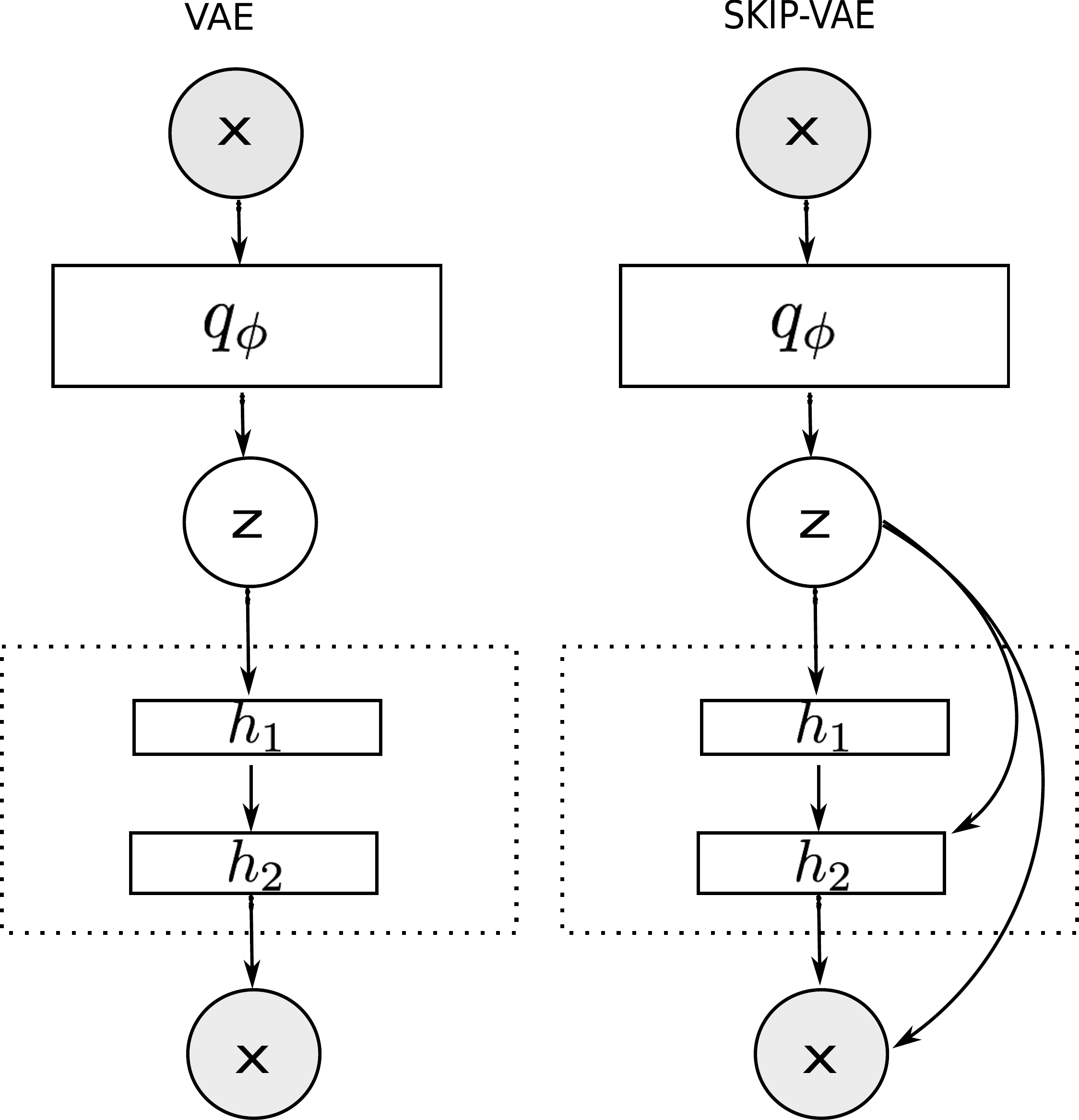}}\quad\quad\quad\quad\quad
  {\includegraphics[scale=0.5]{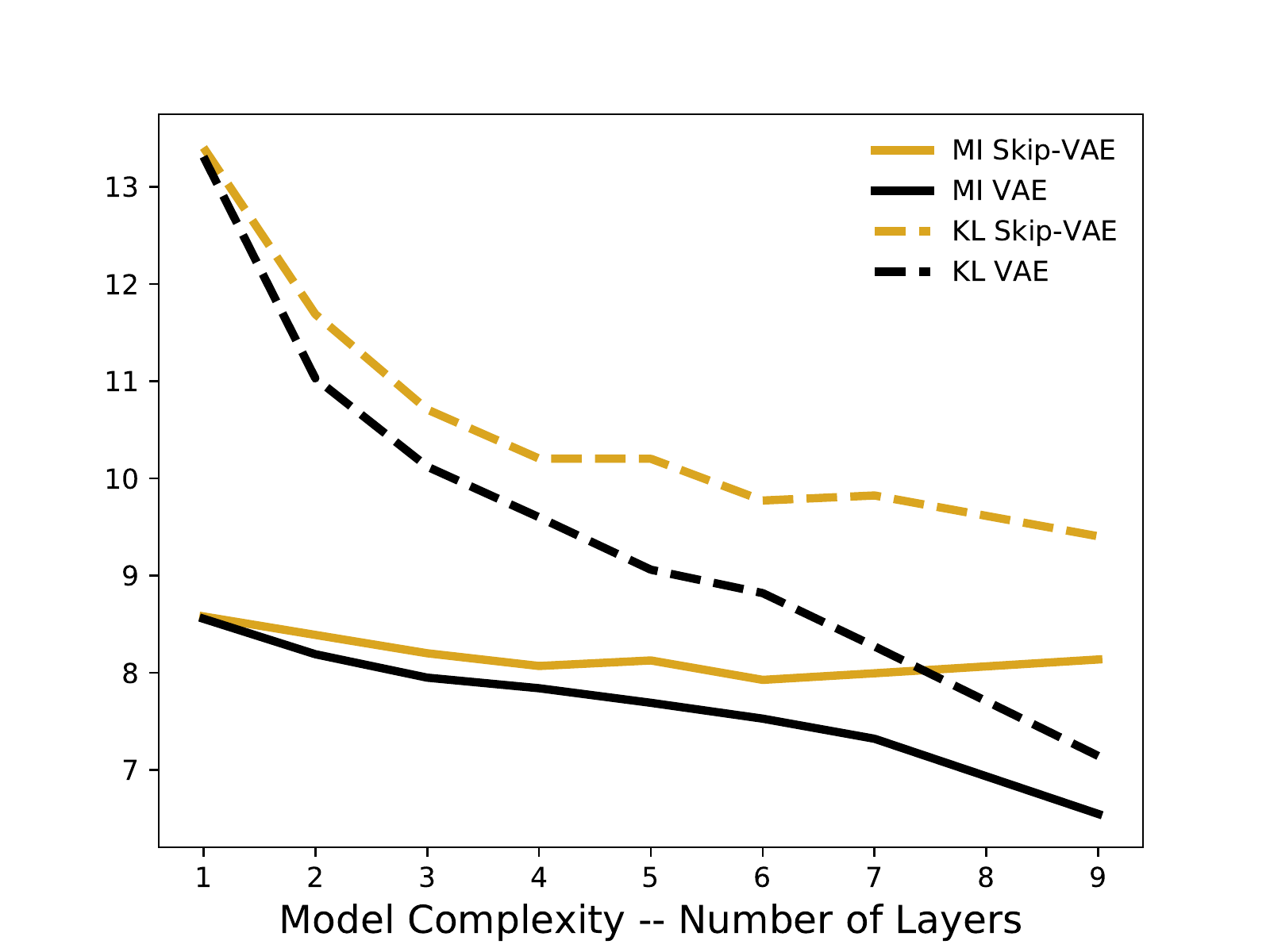}}
  \caption{\textbf{Left:} The \gls{VAE} and \gls{skipVAE} with a two-layer generative model. 
The function $q_{\phi}$ denotes the variational neural network (identical for  \gls{VAE} and \gls{skipVAE}). The difference is in the generative model class: the \gls{skipVAE}'s generative model enforces 
residual paths to the latents at each layer. \textbf{Right:} The mutual information induced by the variational distribution and $\kl$ from the variational distribution to the prior for the \gls{VAE} and the \gls{skipVAE} on MNIST as we vary the number of layers $L$. The \gls{skipVAE} leads to both higher $\kl$ and higher mutual information.}
\label{fig:model}
\end{figure*} 
\section{Latent variable collapse issue in
  \acrshort{VAE}s}\label{sec:collapse}

As we said, the \gls{VAE} binds a deep generative model and
amortized variational inference.  The deep model generates data
through the following process.  First draw a latent variable
$\mbz$ from a prior $p(\mbz)$; then draw an observation $\mbx$ is
$p_{\theta}(\mbx \g \mbz)$, the conditional distribution of
$\mbx$ given $\mbz$.  This likelihood is an exponential family
distribution parameterized by a deep neural network,
\begin{align*}
  p_{\theta}(\mbx \g \mbz)
  &= \text{ExpFam}\left(\mbx; \eta\left(\mbz; \theta\right)\right)\\
   &= \nu(\mbx) \exp\left\{\eta\left(\mbz; \theta\right)^\top\mbx - A\left(\eta\left(\mbz; \theta\right)\right)
    \right\},
\end{align*}
where $A(\cdot)$ is the log-normalizer of the exponential family. The exponential
family provides a compact notation for many types of data, e.g.,
real-valued, count, binary, and categorical. We use this notation to highlight that
\glspl{VAE} can model many types of data.

The natural parameter $\eta(\mbz; \theta)$ is a hierarchical function
of $\mbz$; see Figure\nobreakspace \ref {fig:model} (left). Consider a function with $L$ layers,
where $\mbh^{(l)}$ is the hidden state in the $l^{th}$ layer and
$\mbh^{(1)}$ is the hidden state closest to $\mbz$. The natural
parameter $\eta(\mbz; \theta)$ is computed as follows:
\begin{enumerate}
  \item $\mbh^{(1)} = f_{\theta_0}(\mbz)$ 
  \item $\mbh^{(l+1)} = f_{\theta_{l}}\left(\mbh^{(l)}\right) \quad l = 1 \dots L-1$ 
  \item $\eta(\mbz; \theta) = f_{\theta_L}\left(\mbh^{(L)}\right)$.
\end{enumerate}
The parameter $\theta$ is the collection
$\{\theta_0, \dots, \theta_L\}$. Given data, it should ideally
be fit to maximize the log marginal likelihood; see
Eq.\nobreakspace \ref {eq:marginal-likelihood}.

However the integrals in Eq.\nobreakspace \ref {eq:marginal-likelihood} are intractable. To
circumvent this issue, \glspl{VAE} maximize a lower bound of the log
marginal likelihood, also known as the \gls{ELBO}; see
Eq.\nobreakspace \ref {eq:elbo1}. In the \gls{ELBO}, $q(\mbz \g \mbx; \phi)$ is an
amortized variational distribution; its parameters are fit so that it
approximates the intractable posterior $p(\mbz \g \mbx; \theta)$. The
\gls{ELBO} is tight when
$q(\mbz \g \mbx; \phi) = p(\mbz \g \mbx; \theta)$. The objective
targets both a good likelihood and a good approximate posterior
distribution.

Unfortunately, if the likelihood $p_\theta(\mbx \g \mbz)$ is
too flexible (e.g. a recurrent neural network that fully conditions 
on all previous tokens), it is difficult to achieve this balance. 
Consider an equivalent expression
for the \gls{ELBO},
\begin{align}
  \label{eq:elbo2}
  \gls{ELBO} &= E_{p(\mbx)}
               E_{q_{\phi}(\mbz \g \mbx)} \left[
               \log p_{\theta}(\mbx \g \mbz)
               \right] \nonumber\\
               &- 
               \text{KL}\left(q_{\phi}(\mbz \g \mbx) \gg p(\mbz)\right)
               ,
\end{align}
where $p(\mbx)$ is the population distribution of $\mbx$.  The
flexible likelihood $p_\theta(\mbx \g \mbz)$ in the first term allows the \gls{VAE} to push the
KL term to zero (i.e. setting $  \text{KL}\left(q_{\phi}(\mbz \g \mbx) \gg p(\mbz)\right) \approx 0$) while still giving high probability to the data. This
behavior results in a generative model that gives a meaningless
approximate posteriors and thus poor latent representations.  
\cite{chen2016variational} theoretically justify latent variable collapse
via a ``bits-back" argument: if the likelihood
model is flexible enough to model the data distribution $p(\mbx)$ without using any information
from $\mbz$, then the global optimum is indeed obtained by setting $p_\theta(\mbx \g \mbz) = p(\mbx)$
and $q_{\phi}(\mbz \g \mbx) = p(\mbz)$. 

Let's understand this phenomenon from another angle, which will motivate 
our use of generative skip models. First we define the
\textit{variational joint} distribution.
\begin{definition}\label{def:varjoint}
  For any data $\mbx$ and variational posterior
  $q_{\phi}(\mbz \g \mbx)$, the variational joint $q_{\phi}(\mbx , \mbz)$ is the joint
  distribution of $\mbx$ and $\mbz$ induced by
  $q_{\phi}(\mbz \g \mbx)$. It induces a marginal $q_{\phi}(\mbz)$ called 
  the aggregated posterior~\citep{makhzani2015adversarial, mescheder2017adversarial}
\begin{align*}
q_{\phi}(\mbx , \mbz) &= p(\mbx)\cdot q_{\phi}(\mbz \g \mbx)
\quad \text{and} \quad
q_{\phi}(\mbz) = E_{p(\mbx)}q_{\phi}(\mbz \g \mbx).
\end{align*}
\end{definition}
With this definition in hand, consider a third form of the \gls{ELBO}~\citep{hoffman2016elbo},
\begin{align}\label{eq:elbo3}
  \gls{ELBO} &= E_{p(\mbx)}\left\{ 
               E_{q_{\phi}(\mbz \g \mbx)} \left[
               \log p_{\theta}(\mbx \g \mbz)\right]\right\} \nonumber \\
               &- \mathcal{I}_q(\mbx, \mbz) - \text{KL}(q_{\phi}(\mbz) \gg p(\mbz))
\end{align}
We expressed the $\text{KL}\left(q_{\phi}(\mbz \g \mbx) \gg p(\mbz)\right)$ of Eq.\nobreakspace \ref {eq:elbo2}
as a function of a mutual information
\begin{align}\label{eq:kl}
 \text{KL}\left(q_{\phi}(\mbz \g \mbx) \gg p(\mbz)\right)) &=
       \mathcal{I}_q(\mbx, \mbz) + \text{KL}(q_{\phi}(\mbz) \gg p(\mbz))
    ,
\end{align}
where $ \mathcal{I}_q(\mbx, \mbz)$ is defined as
\begin{align*}
  \mathcal{I}_q(\mbx, \mbz) &=
      E_{p(\mbx)}E_{q_{\phi}(\mbz \g \mbx)}\log q_{\phi}(\mbz \g \mbx) - 
      E_{q_{\phi}(\mbz)} \log q_{\phi}(\mbz) 
      .
\end{align*}
It is the mutual information between $\mbx$ and $\mbz$ induced by the
variational joint and the aggregated posterior.

The \gls{ELBO} in Eq.\nobreakspace \ref {eq:elbo3} reveals that setting the KL term to zero
is equivalent to setting
\begin{align*}
\mathcal{I}_q(\mbx, \mbz) &= \kl(q_{\phi}(\mbz) \gg p(\mbz)) = 0.
\end{align*}
This is true by non-negativity of $\kl$ and mutual information. If $\mbz$
is a good representation of $\mbx$, then the mutual information will be high and
thus the $\kl$ term will be nonzero. But as can be seen 
from Eq.\nobreakspace \ref {eq:elbo3}, the \gls{ELBO} objective contains the negative of the mutual 
information between $\mbz$ and $\mbx$, and thus high mutual information is in contention 
with maximizing the \gls{ELBO}. Of course,  our goal is not merely to prevent the $\kl$ 
from being zero---a trivial way to prevent the $\kl$ from being zero is by only maximizing
\begin{align*}
	\mathcal{L} &=  E_{q_{\phi}(\mbz \g \mbx)} \left[
               \log p_{\theta}(\mbx \g \mbz)\right].
\end{align*}
which essentially corresponds to an auto-encoding objective.
However maximizing this objective leads to poor generative models since the variational distribution 
is unregularized and distinct from the prior---the distribution used to generate samples once training 
is finished. 

We next propose a method that still optimizes the \gls{ELBO} but prevents the $\kl$ from 
collapsing to zero.

\section{Generative skip models avoid latent variable collapse}\label{sec:model}
\begin{figure}[t]
\center
\includegraphics[width=0.49\linewidth, height=4.5cm]{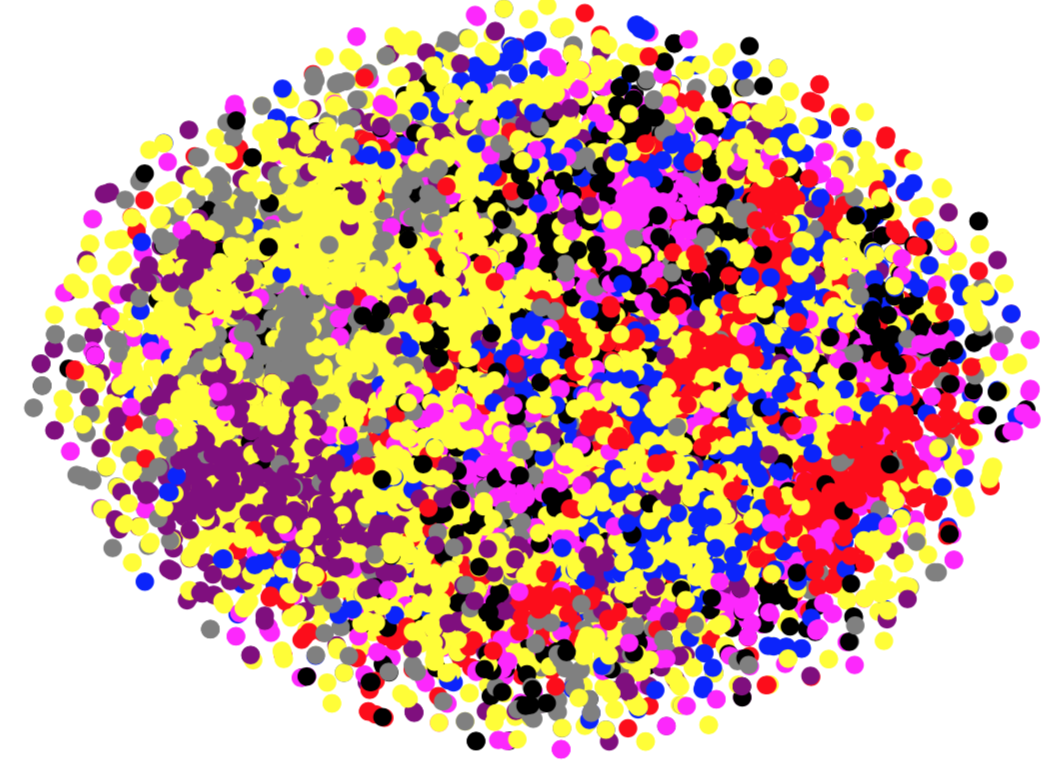}
\includegraphics[width=0.49\linewidth, height=4.5cm]{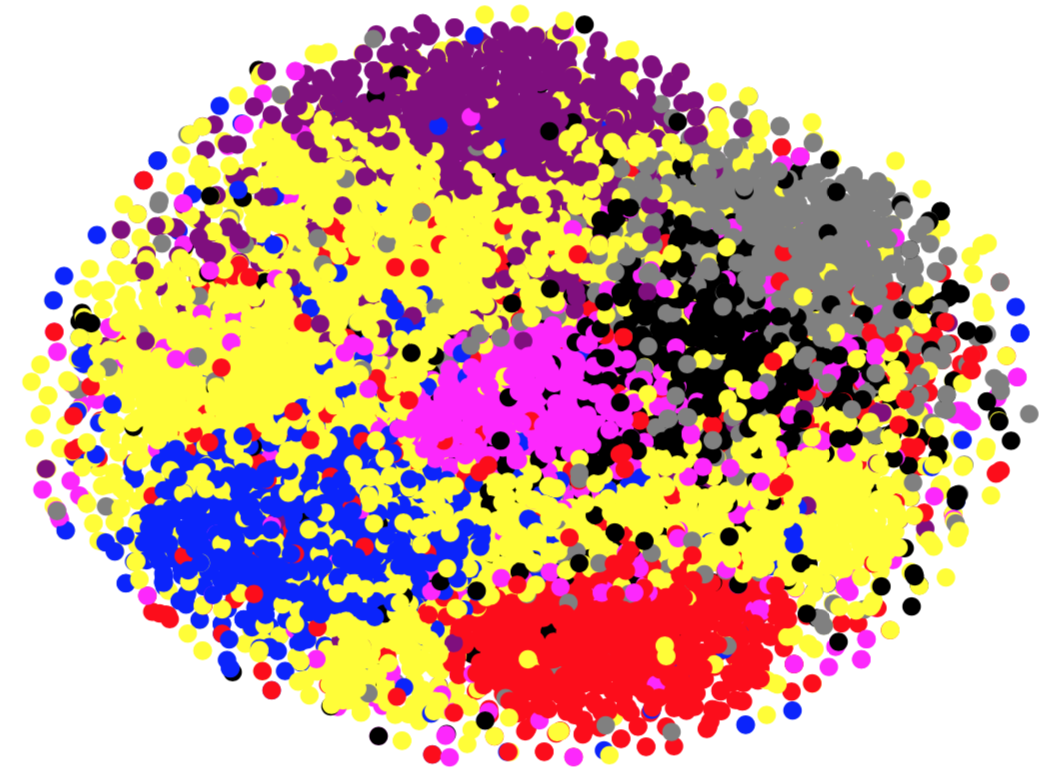}
\caption{Clustering of the latent variables learned by fitting a \gls{VAE} (left) 
and a \gls{skipVAE} (right) on MNIST and applying T-SNE on the test set. 
The model is a 9-layer PixelCNN and the variational neural network is a 3-layer ResNet. 
The colors represent digit labels. The \gls{skipVAE} clusters the latent variables 
better than the \gls{VAE}; it discovers $7$ digit classes. The remaining 3 classes 
are covered by the other classes. The latent variables learned by the \gls{VAE} 
are not meaningful as they are spread out. The \gls{skipVAE} learns more useful latent representations.}
\label{fig:cluster}
\end{figure}

We now describe \glspl{skipVAE}, a family of deep generative 
models that extend \glspl{VAE} to promote high mutual information 
between observations and their associated latent variables. 

A \gls{skipVAE}
is a modified version of the exponential family model described in Section\nobreakspace \ref {sec:collapse}. 
The natural parameter $\eta(\mbz; \theta)$ is now computed as:
\begin{enumerate}
    \item $\mbh^{(1)} = f_{\theta_0}(\mbz)$
    \item $\mbh^{(l+1)} = g_{W_l}\left(f_{\theta_{l}}\left(\mbh^{(l)}\right), \mbz\right)$ for $l = 1 \dots L-1$
    \item $\eta(\mbz; \theta) = g_{W_L}\left(f_{\theta_L}\left(\mbh^{(L)}\right), \mbz\right)$
.
\end{enumerate}
At each layer $l$ of the neural network producing $\eta(\mbz; \theta)$, the hidden state $\mbh^{(l)}$ is a function of
 the latent variable $\mbz$ as well as the previous 
hidden state. The functions $f_{\theta_l}$ are the same 
as in the non-skip generative model; the separately parameterized \textit{skip} 
functions $g_{W_l}$ are nonlinear and combine $\mbz$ with the 
previous hidden state. Figure\nobreakspace \ref {fig:model} (left) illustrates this process.

 Any type of \gls{VAE} can be turned into its \gls{skipVAE} counterpart by adding skips/residual paths to its generative model. 

Our main result is that \glspl{skipVAE} promote higher mutual information between $\mbx$ and $\mbz$ 
when trained with the \gls{ELBO} objective in Eq.\nobreakspace \ref {eq:elbo1}. 

\Glspl{skipVAE} are amenable to any type of skip function $g$; in this section we consider 
 a simple subclass that empirically works well, specifically,
\begin{align*}
g_{W_l}\left(f_{\theta_l}(\mbh^{(l)}), \mbz\right) &=
\sigma\left(
W_l^{(h)}f_{\theta_l}(\mbh^{(l)}) + W_l^{(z)}\mbz
\right)
\end{align*}
where $\sigma$ is a nonlinear function such as sigmoid or ReLU (not used at the last layer), 
and $W_l^{(h)} \neq \mbzero$ and $W_l^{(z)} \neq \mbzero$ are learned deterministic weights. 
Similar to other uses of skip connections~\citep{fukushima1988neocognitron,he2016identity,srivastava2015highway,bahdanau2014neural} 
we do not need to explicitly enforce the constraints $W_l^{(h)} \neq \mbzero$ and $W_l^{(z)} \neq \mbzero$ in practice. 

\begin{thm}\label{tm:skip_theorem}
Consider observation $\mbx$ from an L-layer deep generative model. Denote 
by $\mathcal{I}_p^{\gls{skipVAE}}(\mbx, \mbz)$ the mutual information 
between $\mbx$ and $\mbz$ induced by the generative skip model. Similarly 
denote by $\mathcal{I}_p^{\gls{VAE}}(\mbx, \mbz)$ the mutual information 
between $\mbx$ and $\mbz$ induced by the counterpart generative model 
(the one constructed from the generative skip model by taking the skip connections out). 
Then 
\begin{align*}
		\mathcal{I}_p^{\gls{skipVAE}}(\mbx, \mbz) \geq \mathcal{I}_p^{\gls{VAE}}(\mbx, \mbz) 
\end{align*}
\end{thm}

\parhead{Proof Sketch.}~The proof is based on several applications of the data processing 
inequality of information theory~\citep{cover2012elements} which states that the dependence of $\mbx$ to any hidden state in the hierarchy becomes weaker as 
one moves further away from $\mbx$ in that hierarchy. As a result $\mbx$ will depend 
less on the lower layers than the layers near it. Applying this inequality first in 
the generative skip model yields 
$\mathcal{I}_p^{\gls{skipVAE}}(\mbx, \mbz) \geq \mathcal{I}_p^{\gls{skipVAE}}(\mbx, \mbh^{(l)})$ $\forall l \in \{1, ... , L\}$. 
In particular $\mathcal{I}_p^{\gls{skipVAE}}(\mbx, \mbz) \geq \mathcal{I}_p^{\gls{VAE}}(\mbx, \mbh^{(1)})$. 
Applying the inequality again---this time in the generative model of the \gls{VAE}---we 
have $\mathcal{I}_p^{\gls{VAE}}(\mbx, \mbh^{(1)}) \geq \mathcal{I}_p^{\gls{VAE}}(\mbx, \mbz)$. 
It then follows that $\mathcal{I}_p^{\gls{skipVAE}}(\mbx, \mbz) \geq \mathcal{I}_p^{\gls{VAE}}(\mbx, \mbz)$.

\protect \MakeUppercase {T}heorem\nobreakspace \ref {tm:skip_theorem} says that for any \gls{VAE} one can find a \gls{skipVAE} with higher mutual information. We now use this result to derive the implicit objective function optimized by a \gls{skipVAE}. 

Consider a \gls{VAE} with mutual information $\delta = \mathcal{I}_{\gls{VAE}}(\mbx, \mbz)$. We aim to learn the corresponding \gls{skipVAE}.
Using the result of \protect \MakeUppercase {T}heorem\nobreakspace \ref {tm:skip_theorem}, rewrite the \gls{ELBO} maximization problem 
under the generative skip model as a constrained maximization problem,
\begin{align*}
\theta^*, \phi^* &= 
\arg\max_{\theta, \phi} \gls{ELBO} \quad \text{s.t}  \quad
\mathcal{I}_p^{\gls{skipVAE}}(\mbx, \mbz) \geq \delta
.
\end{align*}
The equivalent Lagrange dual maximizes 
\begin{align*}
\tilde{\mathcal{L}} &= 
\gls{ELBO} + \lambda \mathcal{I}_p^{\gls{skipVAE}}(\mbx, \mbz)
\end{align*}
where $\lambda > 0$ is the corresponding Lagrange multiplier. Using the 
expression of the \gls{ELBO} in Eq.\nobreakspace \ref {eq:elbo1} and using the 
variational joint as defined in Definition\nobreakspace \ref {def:varjoint}, write the objective of \gls{skipVAE} 
as 
\begin{align}\label{eq:lagrange}
\mathcal{L} 
&=\mathcal{H}_p(\mbx) - \lambda \mathcal{I}_p^{\gls{skipVAE}}(\mbx, \mbz) + 
\kl(q_{\phi}(\mbx, \mbz) \gg p_{\theta}(\mbx, \mbz))
,
\end{align}
where $\mathcal{H}_p(\mbx)$ is the entropy of the data distribution. 

Minimizing Eq.\nobreakspace \ref {eq:lagrange} with respect to $\theta$ and $\phi$ is equivalent to joint distribution 
matching\footnote{Joint distribution matching in the context of \glspl{VAE} means making the model joint $p_{\theta}(\mbx, \mbz) = p_{\theta}(\mbx \g \mbz)p(\mbz)$ close to the variational joint $q_{\phi}(\mbx, \mbz) = q_{\phi}(\mbz \g \mbx) p_{\text{data}}(\mbx)$.} under the constraint that the mutual information induced by the 
generative model $ \mathcal{I}_p^{\gls{skipVAE}}(\mbx, \mbz)$ is maximized. 
Minimizing Eq.\nobreakspace \ref {eq:lagrange} brings $p(\mbx, \mbz; \theta)$ 
closer to $q(\mbx, \mbz; \phi)$ thus also increasing $\mathcal{I}_q^{\gls{skipVAE}}(\mbx, \mbz)$---the 
mutual information under the variational joint. Note the \gls{skipVAE} increases $\mathcal{I}_q^{\gls{skipVAE}}(\mbx, \mbz)$ by acting on the generative model to increase $\mathcal{I}_p^{\gls{skipVAE}}(\mbx, \mbz)$. 
In doing so, it mitigates latent variable collapse.
In experiments we see that the \gls{skipVAE} indeed increases $\mathcal{I}_q^{\gls{skipVAE}}(\mbx, \mbz)$ relative to the \gls{VAE}. 
 
\section{Empirical study}\label{sec:empirical}
\begin{table*}[t]
\center
\begin{small}
\caption{Performance of \gls{skipVAE} vs \gls{VAE} on MNIST as the dimensionality 
of the latent variable increases. \gls{skipVAE} outperforms \gls{VAE} on all collapse 
metrics while achieving similar \gls{ELBO} values.}
\begin{tabular}{c  c c c c c c c c c c c}
\toprule
  &  \multicolumn{2}{c}{\gls{ELBO}} & & \multicolumn{2}{c}{ $\kl$ } & & \multicolumn{2}{c}{MI} & & \multicolumn{2}{c}{AU}  \\
Dim &  \gls{VAE} & \gls{skipVAE} & & \gls{VAE} & \gls{skipVAE} & & \gls{VAE} & \gls{skipVAE} & & \gls{VAE} & \gls{skipVAE}\\
\midrule
$2$ & $\textbf{-84.27}$ & $\text{-}84.30$ & &$3.13$ & $\textbf{3.54}$  &  & $3.09$ & $\textbf{3.46}$ & & $2$ & $2$\\
$10$ & $\text{-}83.01$ &  $\textbf{-82.87}$ & & $8.29$ & $\textbf{9.41}$ & & $7.35$ & $\textbf{7.81}$ & & $9$ &  $\textbf{10}$\\
$20$ & $\text{-}83.06$ &  $\textbf{-82.55}$ & &$7.14$ & $\textbf{9.33}$    &  & $6.55$ & $\textbf{7.80}$  & & $8$ &  $\textbf{13}$\\
$50$ & $\text{-}83.31$ & $\textbf{-82.58}$ & &$6.22$ & $\textbf{8.67}$  &  & $5.81$ & $\textbf{7.49}$ & & $8$ & $\textbf{12}$\\
$100$ & $\text{-}83.41$ &  $\textbf{-82.52}$ & & $5.82$ & $\textbf{8.45}$ & & $5.53$ & $\textbf{7.38}$ & & $5$ &  $\textbf{9}$\\
\bottomrule
\end{tabular}\label{tab:mnist1}
\end{small}
\end{table*}

\begin{table*}[t]
\center
\begin{small}
\caption{Performance of \gls{skipVAE} vs \gls{VAE}  on MNIST (Top) and 
Omniglot (Bottom) as the complexity of the generative model increases. The number of latent dimension is fixed at $20$. Skip-VAE 
outperforms VAE on all collapse metrics while achieving similar \gls{ELBO} values, and the difference widens as layers increase.}
\begin{tabular}{c c c c c c c c c c c c c}
\toprule
&  &  \multicolumn{2}{c}{\gls{ELBO}} & & \multicolumn{2}{c}{ $\kl$ } & & \multicolumn{2}{c}{MI} & & \multicolumn{2}{c}{AU}  \\
& Layers & \gls{VAE} & \gls{skipVAE} & & \gls{VAE} & \gls{skipVAE} & & \gls{VAE} & \gls{skipVAE} & & \gls{VAE} & \gls{skipVAE}\\
\midrule
\multirow{4}{*}{MNIST} & $1$ & $\text{-}89.64$ & \textbf{-89.22} & &$13.31$ & $\textbf{13.40}$  &  & $8.56$ & $8.56$ & & $20$ & $20$\\
&$3$ & $\text{-}84.38$ &  $\textbf{-84.03}$ & & $10.12$ & $\textbf{10.71}$ & & $7.95$ & $\textbf{8.20}$ & & $16$ &  $16$\\
&$6$ & $\text{-}83.19$ &  $\textbf{-82.81}$ & &$8.82$ & $\textbf{9.77}$    &  & $7.53$ & $\textbf{7.93}$  & & $11$ &  $\textbf{13}$\\
&$9$ & $\text{-}83.06$ & $\textbf{-82.55}$ & & $7.14$ &  $\textbf{9.34}$   & & $6.55$ & $\textbf{7.80}$ & & $8$ &  $\textbf{13}$\\
\midrule
\multirow{4}{*}{Omniglot} &$1$ & $\text{-}97.69$ & $\textbf{-97.66}$  & & $\textbf{8.42}$ & $8.37$  &  & $\textbf{7.09}$ & $7.08$  & &$20$ & $20$\\
&$3$ & $\text{-}93.95$ & $\textbf{-93.75}$ & &$6.43$  & $\textbf{6.58}$ &  & $5.88$ & $\textbf{5.97}$ & &$20$ &   $20$\\
&$6$ & $\text{-}93.23$ & $\textbf{-92.94}$  & &$5.24$ & $\textbf{5.78}$ &  & $4.94$ & $\textbf{5.43}$ & &$20$ &  $20$\\
&$9$ &$\text{-}92.79$ & $\textbf{-92.61}$ & & $4.41$ & $\textbf{6.12}$ & & $4.24$ & $\textbf{5.65}$  & & $11$ &  $\textbf{20}$\\
\bottomrule
\end{tabular}\label{tab:mnist2}
\end{small}
\end{table*}

\begin{table*}[t]
\center
\begin{small}
\caption{\gls{VAE} and \gls{skipVAE} on MNIST using $50$ latent dimensions with a simplified network. Here the encoder is a 2-layer MLP with 512 units in each layer and the decoder is also an MLP. The results below correspond to different number of layers for the decoder.}
\begin{tabular}{c c c c c c c c c c c c}
\toprule
  &  \multicolumn{2}{c}{\acrshort{ELBO}} & & \multicolumn{2}{c}{ $\kl$ } & & \multicolumn{2}{c}{MI} & & \multicolumn{2}{c}{AU}  \\
Layers & \acrshort{VAE} & \acrshort{skipVAE} & & \acrshort{VAE} & \acrshort{skipVAE} & & \acrshort{VAE} & \acrshort{skipVAE} & & \acrshort{VAE} & \acrshort{skipVAE}\\
\midrule
$2$ & $-94.88$ &  $\textbf{-94.80}$ & & $24.23$ & $\textbf{26.35}$ & & $\textbf{9.21}$ & $9.20$ & & $17$ &  $\textbf{24}$\\
$3$ & $-95.38$ &  $\textbf{-94.17}$ & & $21.87$ & $\textbf{26.15}$ & & $9.20$ & $\textbf{9.21}$ & & $13$ &  $\textbf{21}$\\
$4$ & $-97.09$ &  $\textbf{-93.79}$ & & $20.95$ & $\textbf{25.63}$ & & $\textbf{9.21}$ & $\textbf{9.21}$ & & $11$ &  $\textbf{21}$\\
\bottomrule
\end{tabular}\label{tab:mnist3}
\end{small}
\end{table*}

\begin{table*}[t]
\center
\begin{small}
\caption{\gls{skipVAE} and \textsc{skip-sa-vae} perform better than their 
counterparts (\gls{VAE},  \gls{SAVAE}) on the Yahoo corpus under all latent 
variable collapse metrics while achieving similar log-likelihoods. In particular, all 
latent dimensions are active when using \textsc{skip-sa-vae}. Perplexity (PPL) for 
the variational models  is estimated by importance sampling of the log marginal likelihood with 
200 samples from $q_{\phi}(\zvec \g \mbx)$. }
\begin{tabular}{l@{\hskip 0.5cm} c@{\hskip 1.5cm} c@{\hskip 1cm} c@{\hskip 1cm} c@{\hskip 1cm} c@{\hskip 1cm}  c  }
\toprule
Model &Dim& PPL & $\gls{ELBO}$ & $\kl$ &  MI & AU  \\
\midrule
\textsc{lstm language model} &- & $61.60$ & - & - & - & -\\
\hline
\gls{VAE}     &$32$ & $ 62.38$  & $-330.1$ & $0.005$ & $0.002$ & $0$  \\
\gls{skipVAE} &$32$ & $ 61.71$  & $-330.5$ & $\textbf{0.34}$ & $\textbf{0.31}$ & $\textbf{1}$  \\
\midrule
 \gls{SAVAE}  & $32$&$ 59.85$ & $-327.5$ & $5.47$ & $4.98$ & $14$ \\
\textsc{skip-sa-vae}  &$32$ & $ 60.87$ &$-330.3$ & $\textbf{15.05}$ & $\textbf{7.47}$ & $\textbf{32}$   \\
\midrule
\gls{SAVAE}   & $64$&  $60.20$ & $-327.3$ & $3.09$ & $2.95$ & $10$ \\
\textsc{skip-sa-vae} &$64$ & $ 60.55$ & $-330.8$ & $\textbf{22.54}$ & $\textbf{9.15}$ & $\textbf{64}$  \\
\bottomrule
\end{tabular}\label{tab:yahoo}
\end{small}
\end{table*}

We assess \glspl{skipVAE} by applying skip connections to extend a standard \gls{VAE}~\citep{kingma2013auto,rezende2014stochastic} 
and to the recently introduced \gls{SAVAE}~\citep{kim2018semi}. We use benchmark datasets for images and text: MNIST, Omniglot, and the Yahoo corpus. Text datasets have been shown to be particularly sensitive to latent variable collapse when the likelihood is parameterized as a fully autoregressive model, such as a recurrent neural network \citep{bowman2015generating}. Note that we are interested in learning both a good generative model (as measured by
the ELBO) and a good latent representation of the data (as measured by mutual information
and other metrics). The prior for all studies is a spherical Gaussian, and the variational posterior is a diagonal Gaussian. We compare the performance of \gls{skipVAE} and the baselines when varying the dimensionality of the latent variable and the complexity of the generative model.

\paragraph{Evaluation} 
We assess predictive performance---as given by a measure of held-out log-likelihood---and latent variable collapse. For image datasets we report the \gls{ELBO} as a measure of log-likelihood; for text we report both the \gls{ELBO} and perplexity (estimated using importance sampling). 

Assessing latent variable collapse is more difficult. We employ three metrics: the KL-divergence, mutual information (MI), and number of active units (AU).

The first metric is the $\kl$ regularization term of the \gls{ELBO} as written in Eq.\nobreakspace \ref {eq:elbo1}. 

The second measure is the mutual information induced by the variational joint $\mathcal{I}_q(\xvec, \zvec)$. 
Using the expression of the $\kl$ in Eq.\nobreakspace \ref {eq:kl} we have
\begin{align*}
	\mathcal{I}_q(\xvec, \zvec) &=  
		\text{KL}\left(q_{\phi}(\mbz \g \mbx) \gg p(\mbz)\right)) - \text{KL}(q_{\phi}(\mbz) \gg p(\mbz)).
\end{align*}
We follow \cite{hoffman2016elbo} and approximate this mutual information using Monte Carlo estimates of the two $\kl$ terms. In particular,
\begin{align*}
\kl(q_\phi(\zvec) \, \Vert \, p(\zvec)) &= 
\mathbb{E}_{q_\phi(\zvec)} \left[ \log q_\phi(\zvec) - \log p(\zvec)\right] \\
&\approx 
\frac{1}{S}\sum_{s=1}^{S}
\log q_\phi(\mbz^{(s)}) - \log p(\mbz^{(s)})
\end{align*} 
where each aggregated posterior $q_\phi(\mbz^{(s)})$ is also approximated by Monte Carlo.

The third measure of latent variable collapse is the number of "active" units of 
the latent variable $\mbz$. This is defined in \cite{Burda2015} as
\begin{align*}
\text{AU} &= \sum_{d=1}^{D} \mathbbm{1}{\left\{\text{Cov}_{p(\xvec)}\left(\mathbb{E}_{q_{ \phi}(\mbz \g \xvec)}[\mbz_d]
\right) \geq \epsilon
\right\}},
\end{align*}
where $z_d$ is the $d^{th}$ dimension of $\mbz$ and $\epsilon$ is a 
threshold. ($\mathbbm{1}\{\cdot\}$ is an indicator giving $1$ when 
its argument is true and $0$ otherwise.)
We follow \cite{Burda2015} and use a 
threshold of $\epsilon = 0.01$. We observe the same phenomenon: the 
histogram of the number of active dimensions of $\mbz$ is bi-modal, which 
means that it is not highly sensitive to the chosen threshold.

\subsection{Images}
\paragraph{Model} We use a 3-layer ResNet \citep{He2016} (with $3\times 3$ filters and $64$ feature maps in 
each layer) as the variational neural network and a 9-layer Gated 
PixelCNN \citep{Oord2016b} (with $3 \times 3$ filters and $32$ feature maps) as the likelihood. The baseline model uses a linear map from the sample (to project out to the image spatial resolution), concatenated with the original image, which is fed to the PixelCNN. This setup reflects the current state-of-the-art for image \glspl{VAE}~\citep{gulrajani2016pixelvae,chen2016variational}.\footnote{While our model capacity is similar to these works, our performance is slightly worse since we do not employ additional techniques such as
data-dependent initialization \citep{chen2016variational}.} The \gls{skipVAE} uses a linear map from the sample, concatenated with the output from each layer of the PixelCNN (before feeding it to the next layer). While this results in slightly more parameters for the \gls{skipVAE} model, we found that the baseline \gls{VAE}'s performance on the collapse metrics actually gets worse as the model size increases. 

\paragraph{Results}
Table\nobreakspace \ref {tab:mnist1} shows the results on MNIST as we vary the size of the latent dimension. In all scenarios, the generative skip model yields higher $\kl$ between the variational posterior and the prior, higher mutual information (confirming the statement in \protect \MakeUppercase {T}heorem\nobreakspace \ref {tm:skip_theorem}), and uses more latent dimensions (as measured by AU). 

Table\nobreakspace \ref {tab:mnist2} shows experiments on both MNIST and Omniglot as we vary the generative model's complexity by increasing its depth. We use a model with $20$-dimensional latent variables. For \gls{VAE}, as the generative model becomes more expressive the model becomes less reliant on $\zvec$. We see this in the poor performance on the collapse metrics. The \gls{skipVAE} mitigates this issue and performs better on all latent-variable collapse metrics. Note the \gls{ELBO} is similar for both models. These results indicate that the family of generative skip models has a strong inductive bias to share more mutual information between the observation and the latent variable.

Similar results are observed when using weaker models. For example in Table\nobreakspace \ref {tab:mnist3} 
we used \glspl{MLP} for both the variational neural network and the generative model,  and we set the dimensionality of the latent variables to $50$. Even with this weaker setting
the \gls{skipVAE} leads to less collapse than the \gls{VAE}. 

\parhead{Latent Representations.}
We find qualitatively that the latent representations learned by the \gls{skipVAE}
better capture the underlying structure. Figure\nobreakspace \ref {fig:cluster} illustrates this. It shows a much clearer separation of the MNIST digits with the latent space learned by the \gls{skipVAE} compared to the latent space of the \gls{VAE}. \footnote{Note we did not fit a \gls{VAE} and a \gls{skipVAE} with 2-dimensional latents for the visualization. Fitting 2-dimensional latents would have led to much better learned representations for both the \gls{VAE} and the \gls{skipVAE}. However using 2-dimensional latents does not correspond to a realistic setting in practice. Instead we fit the \gls{VAE} and the \gls{skipVAE} on $50$-dimensional latents---as is usual in state-of-the-art image modeling with \glspl{VAE}---and used t-SNE to project the learned latents on a $2$-dimensional space.} 

Quantitatively we performed a classification study on MNIST
using the latent variables learned by the variational neural networks of \gls{VAE} and \gls{skipVAE} as features. This study uses $50$ latent dimensions, a $9$-layer PixelCNN as the generative model, a $3$-layer ResNet as the variational neural 
network, and a simple $2$-layer \gls{MLP} over the posterior means as the classifier. The \gls{MLP} has $1024$ hidden units, ReLU activations, and a dropout rate of $0.5$. The classification accuracy of the \gls{VAE} is $97.19\%$ which is 
lower than the accuracy of the \gls{skipVAE} which is $98.10\%$. 
We also studied this classification performance on a weaker model. We replaced the $9$-layer PixelCNN and the $3$-layer ResNet above by two \glspl{MLP}. The \gls{VAE} achieved an accuracy of $97.70\%$  whereas the \gls{skipVAE} achieved an accuracy of $98.25\%$. 

\subsection{Text}

\paragraph{Model}

For text modeling, we use the training setup from \cite{kim2018semi}, a strong baseline that outperforms standard LSTM language models. The variational neural network is a $1$-layer LSTM with $1024$ hidden units, whose last hidden state is used to predict the mean vector and the (log) variance vector of the variational posterior. The generative model is also a $1$-layer LSTM with $1024$ hidden units. We found that the $1$-layer model performed better than deeper models, potentially due to overfitting. In the \glspl{VAE} the sample is used to predict the initial hidden state of the decoder and also fed as input at each time step. In the generative skip model we also concatenate the sample with the decoder's hidden state.

We also study the semi-amortized variational autoencoder, \gls{SAVAE} \citep{kim2018semi}, 
which proposes a different optimization-based strategy for targeting the latent variable collapse issue when training \glspl{VAE} for text. \gls{SAVAE} combines stochastic variational inference \citep{Hoffman2013} with amortized variational inference by first using an inference network over $\xvec$ to predict the initial variational parameters and then subsequently running iterative inference on the \gls{ELBO} to refine the initial variational parameters. We used $10$ steps of iterative refinement for \gls{SAVAE} and \acrshort{SkipSAVAE}.

\paragraph{Results}
We analyze the Yahoo Answers dataset from \cite{Yang2017}, a benchmark for deep generative models of text. Table\nobreakspace \ref {tab:yahoo} shows the results. We first note that successfully training standard \glspl{VAE} for text with flexible autoregressive likelihoods such as LSTMs remains a difficult problem. We see that \gls{VAE} by itself experiences latent variables collapse.
The \gls{skipVAE} is slightly better than \gls{VAE} at avoiding latent variable collapse for similar log likelihoods, although the $\kl$ is only marginally above zero and the model only has one active unit.

When combining both approaches with the semi-amortized training, we see better use of latent variables in \gls{SAVAE} and \acrshort{SkipSAVAE}.  While \gls{SAVAE} alone does mitigate collapse to an extent, skip connections learn generative models where the mutual information is even higher. Furthermore, the trend changes when adding more latent dimension. For the vanilla \gls{SAVAE}, the mutual information and active units are actually \emph{lower} for a model trained with $64$-dimensional latent variables than a model trained with $32$-dimensional latent variables. This is a common issue in \glspl{VAE} whereby simply increasing the dimensionality of the latent space actually results in a worse
model. In contrast, models trained with skip connections make full use of the latent space and collapse metrics improve as we increase the number of dimensions. For example the \textsc{skip-sa-vae} uses all the dimensions of the latent variables.

\section{Conclusion}\label{sec:discussion}
We have proposed a method for reducing latent variable collapse in \glspl{VAE}. The approach uses skip connections to promote a stronger dependence between the observations and their associated latent variables. The resulting family of deep generative models (\glspl{skipVAE}) learn useful summaries of data. Theoretically we showed that \glspl{skipVAE} yield higher mutual information than their counterparts. We found that \glspl{skipVAE}---when used with more sophisticated \glspl{VAE} such as the \gls{SAVAE}---lead to a significant improvement in terms of latent variable collapse. 
\section*{Acknowledgements} 
We thank Francisco Ruiz for presenting our poster at the Theoretical Foundations and Applications of Deep Generative Models Workshop at ICML, 2018. DMB is supported by ONR N00014-15-1-2209, ONR 133691-5102004,  NIH 5100481-5500001084, NSF CCF-1740833, the Alfred P. Sloan Foundation,  the John Simon Guggenheim Foundation, Facebook, Amazon, and IBM. AMR is supported by NSF-CCF 1704834, Google, Facebook, Bloomberg, and Amazon. YK is supported by a Google PhD Fellowship. 

\bibliography{main}
\bibliographystyle{abbrvnat}

\end{document}